\documentclass[conference]{IEEEtran}
\IEEEoverridecommandlockouts
\usepackage{cite}
\usepackage{amsmath,amssymb,amsfonts}
\usepackage{graphicx}
\usepackage{textcomp}
\usepackage{xcolor}
\usepackage{subcaption}

\usepackage{times}
\usepackage{latexsym}
\usepackage{multirow} 

\usepackage{booktabs}
\usepackage{multirow}

\usepackage{amssymb,mathrsfs,amsmath}
\usepackage{algorithm}  
\usepackage{algpseudocode}  
\usepackage{amsmath}  
\usepackage{graphicx}
\usepackage{hyperref}

\usepackage{multirow}
\usepackage{booktabs} 
\usepackage{hyperref}
\def\BibTeX{{\rm B\kern-.05em{\sc i\kern-.025em b}\kern-.08em
    T\kern-.1667em\lower.7ex\hbox{E}\kern-.125emX}}
\begin{document}

\title{TPN: Transferable Proto-Learning Network towards Few-shot Document-Level Relation Extraction
\thanks{\dag \quad Corresponding Author}

}

\author{\IEEEauthorblockN{Yu Zhang$^1$, Zhao Kang$^{1,2}$\textsuperscript{\dag}}
	\IEEEauthorblockA{$^1$University of Electronic Science and Technology of China, Chengdu, China\\
 $^2$Kashi Institute of Electronics and Information Industry, Xinjiang, China}
        \IEEEauthorblockA{yuzhang2717@gmail.com, zkang@uestc.edu.cn}}

\maketitle

\begin{abstract}
Few-shot document-level relation extraction suffers from poor performance due to the challenging cross-domain transferability of NOTA (\textit{none-of-the-above}) relation representation. 
In this paper, we introduce a \textbf{T}ransferable \textbf{P}roto-Learning \textbf{N}etwork (TPN) to address the challenging issue.
It comprises three core components: Hybrid Encoder hierarchically encodes semantic content of input text combined with attention information to enhance the relation representations. 
As a plug-and-play module for Out-of-Domain (OOD) Detection, Transferable Proto-Learner computes NOTA prototype through an adaptive learnable block, effectively mitigating NOTA bias across various domains. 
Dynamic Weighting Calibrator detects relation-specific classification confidence, serving as dynamic weights to calibrate the NOTA-dominant loss function. 
Finally, to bolster the model's cross-domain performance, we complement it with virtual adversarial training (VAT). 
We conduct extensive experimental analyses on FREDo and ReFREDo, demonstrating the superiority of TPN. Compared to state-of-the-art methods, our approach achieves competitive performance with approximately half the parameter size. Data and code are available at \href{https://github.com/EchoDreamer/TPN} {https://github.com/EchoDreamer/TPN}.
\end{abstract}


\section{Introduction}
\begin{figure}
\centering
\includegraphics[width=0.9\linewidth]{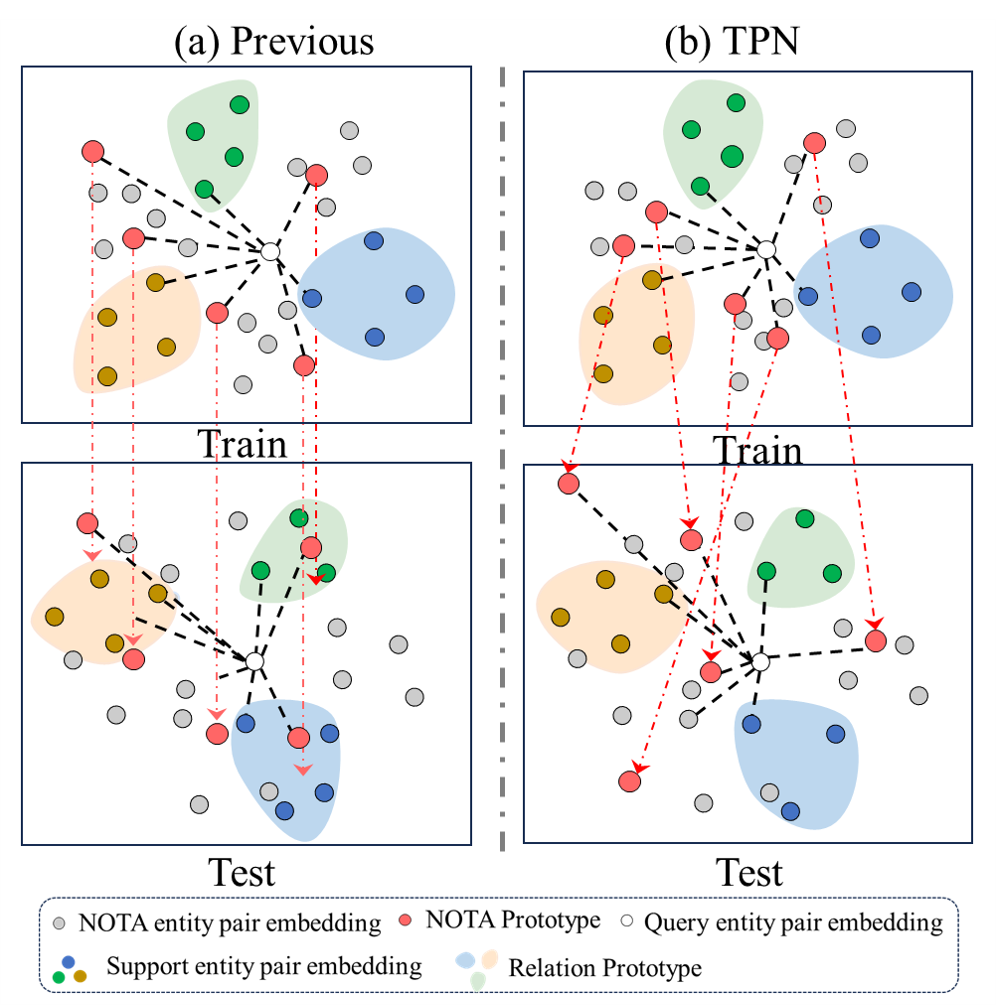}
\caption{Illustration of relation prototype embedding space for previous global multi-proto (left) and our TPN (right). The previous methods share the embedding of the NOTA prototype across different domains. Instead, our TPN dynamically adapts the NOTA prototype from Domain$_{train}$ to Domain$_{test}$.} 
\label{fig:contrast}
\end{figure}

In practical scenarios, document-level relation extraction (DLRE) \cite{docred} has gained prominence due to its effectiveness in extracting complex relations from extensive data. Although numerous efforts \cite{atlop,dlre1} have contributed to effectively improving the performance of DLRE, the method is still constrained by the need for a large amount of annotated data. Inspired by meta learning \cite{meta-learning}, few-shot document-level relation extraction (FSDLRE) \cite{fredo} is proposed to address this issue. But this task suffers from poor performance due to modeling of the NOTA relation. Existing works fail to address cross-domain migratability of NOTA representation. As shown in Fig. \ref{fig:contrast}, previous methods directly model the NOTA prototype using a global multi-proto, which is very limited for cross-domain tasks. Because of the different true relation distributions defined in different domains, the distribution of NOTA varies greatly from domain to domain, which results in a significant degradation of the model generalization performance. 

To tackle the above issue, we propose a novel Transferable Proto-Learning Network (TPN) to adapt to the more complex semantic scenario for FSDLRE. It includes a Hybrid Encoder, a novel plug-and-play Transferable Proto-Learner, and a Dynamic Weighting Calibrator. Hybrid Encoder hierarchically encodes the global and local semantic information of the input document with attention information to enhance the relation representations. Transferable Proto-Learner learns the regularized representation of NOTA relation, which models the way to compute NOTA relation to eliminate NOTA bias in different domains. Dynamic Weighting Calibrator aims to detect relation-specific classification confidence, serving as dynamic weights for different relations to calibrate the NOTA-dominant loss function. Finally, we design virtual adversarial training to encourage the embeddings of the sentence in different domains as closely as possible in the shared latent space. Unlike \cite{9413437}, which utilizes Projected Gradient Descent (PGD) \cite{PGD} to enable us to perform diversified adversarial training on large-scale state-of-the-art models and soft labels to constrain the direction of adversarial samples, we use FreeLB \cite{FreeLB} which performs multiple PGD iterations to craft adversarial examples and uses a hard label (ground truth) to constrain adversarial samples more efficiently.

Compared to the most recent work, RAPL \cite{ReFREDo}, which introduces a relation-aware contrastive learning approach with two independent encoders, our method achieves competitive performance with approximately half the parameter size (123MB VS 221MB) because RAPL uses two separate text encoders, whereas we use only one. Besides, RAPL generates the NOTA prototype by leveraging the hard distance constraints from support relation instances. Instead, as a soft learnable module, our Transferable Proto-Learner will learn more flexiable features about the representation of the NOTA prototype.

The contributions of this paper are summarized as follows: \\
(1) We propose a novel Transferable Proto-Learning Network (TPN) to adapt to the more complex semantic scenario for FSDLRE. It includes a Hybrid Encoder, a novel plug-and-play Transferable Proto-Learner, and a Dynamic Weighting Calibrator. \\
(2) We introduce virtual adversarial training to encourage word embeddings in different domains as closely as possible in the shared latent space.\\
(3) We conduct extensive experimental analysis on FREDo and ReFREDo that demonstrate the superiority of the framework.

\section{Related Work}
\subsection{Relation Extraction}
Advancements in relation extraction (RE), particularly sentence-level relational extraction (SLRE), have seen rapid progress \cite{re-sentence2,fcds,re-sentence1,Document}.

However, in practical scenarios, document-level relation extraction (DLRE) \cite{docred} has gained prominence due to its effectiveness in extracting complex relationships from extensive datasets. 
DLRE, though promising, presents several significant challenges including the identification of relations that span multiple sentences \cite{focal-loss} and imbalance in the distribution of samples, commonly referred to as the long-tail distribution \cite{long-tail}. Attention mechanism is applied to a variety of tasks \cite{tiefake,atlop}. 
ATLOP \cite{atlop} harnessed attention mechanism derived from PLM to enhance the understanding of length textual content and introduced Adaptive Thresholding Loss to tackle the multi-class classification issue. 
Furthermore, SSAN \cite{ssan} incorporates these structural dependencies within the standard self-attention mechanism throughout the overall encoding stage.

\subsection{Few-Shot Sentence-Level Relation Extraction}
Due to the substantial cost associated with manual annotation for relation extraction, Transfer Learning \cite{transfer-learning} emerges as an ideal solution. In the realm of transfer learning, whose essence lies in acquiring the ability to generalize from different datasets \cite{relation-prompt,transfer-learning-ability}, FSRE has gained extensive traction and demonstrated promising performance, for example \cite{few-shot-RE1,few-shot-RE2}. Many FSRE methods have sought to establish meta-learning frameworks based on prototype networks, as pioneered by \cite{NIPS2017_cb8da676}, with the objective of cultivating a class-agnostic metric space. \cite{wang-etal-2022-drk} has taken a different approach by leveraging relation descriptions and knowledge graphs to enhance relation representation, while \cite{label-prompt-dropout} introduced the concept of \textit{Label Prompt Dropout}, which selectively omits label descriptions during the learning process to enhance generalization.

\subsection{Few-Shot Document-Level Relation Extraction}
The application of relation extraction in more complex scenarios is still at an early stage. \cite{fredo} curated the FREDo dataset, a benchmark designed to assess few-shot document-level relation extraction. This benchmark specifically focuses on the prevalent challenges of Long-Tail \cite{long-tail} and Out-of-Distribution Detection \cite{ood} commonly encountered in real-world low-resource scenarios. In RE, OOD targets at \textit{none-of-the-above} (NOTA) distribution. \cite{fredo} utilized the MNAV architecture \cite{sabo-etal-2021-revisiting} as a baseline to analyze the difficulties and research directions. Recently, \cite{ReFREDo} presented a relation-aware prototype learning approach to refine the relation prototypes and generate task-specific NOTA prototype achieving the state-of-the-art.

\begin{figure*}[t]
\centering
\includegraphics[width=0.9\linewidth]{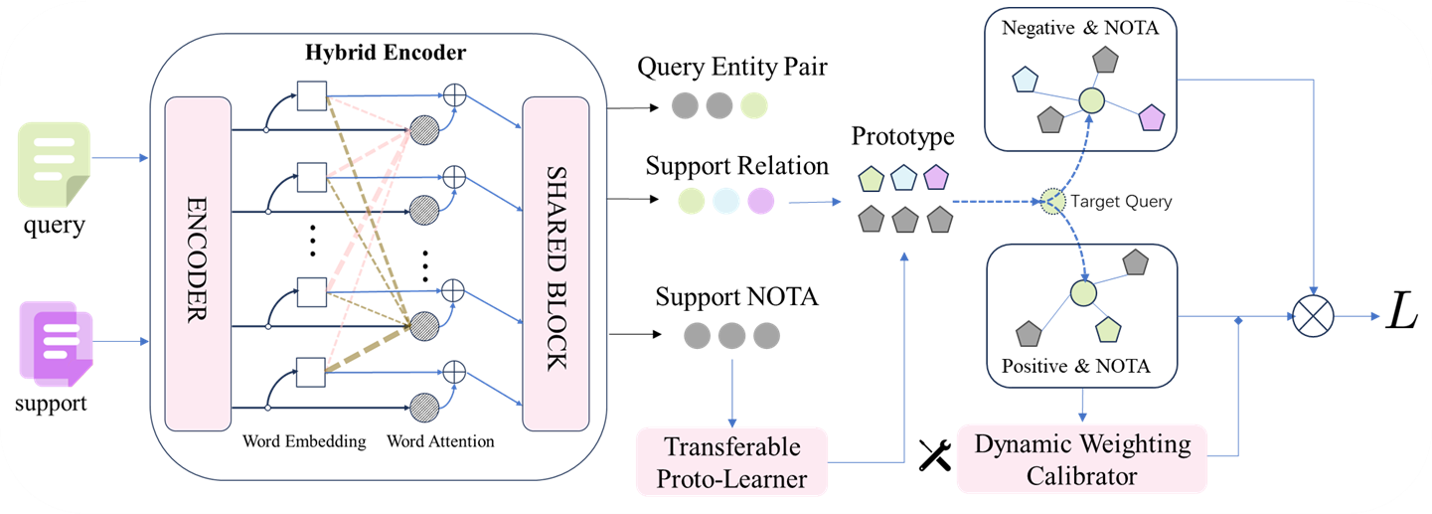}
\caption{The overall architecture of our proposed TPN framework.} 
\label{fig:frame}
\end{figure*}

\section{Problem Formulation}
In this section, we begin by outlining the task definition.
In the FSDLRE task, the entity mentions in each document are pre-annotated. For every training/ testing step, an episode contains a set of support documents $S=\{D_{s_1},D_{s_2},...,D_{s_m}\}$, the corresponding valid triple sets pre-annotated, and a set of query documents $Q=\{D_{q_1},D_{q_2},...,D_{q_m}\}$, where ${s_m}$ and ${q_m}$ represent the number of support document and query document in an episode. In the 3-Doc task, $s_m=3$ and $q_m=1$ and in the 1-Doc task, $s_m=1$ and $q_m=3$.
A triple set in the support document contains all valid triples $(e_h,r_i,e_t)$, where $e_h$ and $e_t$ are the head-entity and tail-entity for a specific relation instance $r_i$ respectively. Note that the annotations of the support documents are complete, meaning that any candidate entity pair without an assigned relation type can be considered NOTA. The goal is to predict all possible valid triples in the query set based on the relation types and semantic information from the support documents.     

Our approach follows the typical meta-learning paradigm \cite{fredo}. In the training phase, we sample an episode containing a subset of relation $R_{epi}$ from $ R_{train}$, where $R_{train}$ is the relation set in the training corpus.
The proto-based model will learn the way to compute the relation prototypes through support documents. In the test phase, the model is evaluated on the test episodes sampled from test corpus containing the set of relations $R_{test}$ which is disjoint with $R_{train}$ to predict all possible valid triples in the query set.

\section{Framework of TPN}
In this section, we introduce the TPN framework. Subsequently, we provide a comprehensive overview of each individual module.
As shown in Fig. \ref{fig:frame}, in each iteration step, the support documents and the query documents are first fed into the Hybrid Encoder to obtain hybrid representations. Then the Transferable Proto-Learner module is employed to obtain the prototype of NOTA relation. Furthermore, we compute the classification confidence to dynamically calibrate the NOTA-dominant loss.

\subsection{Hybrid Encoder}
To enhance the model's contextual relation representation, we propose the Hybrid Encoder to hierarchically encode global and local semantic information of the input text combined with attention information. Given a document $D=\left\{ w_1,w_2,...,w_l \right\}$, we use a pre-trained language model (PLM) to obtain the contextualized embeddings of this document:
\begin{equation}
h_{plm}=PLM\left( \left[ w_1,w_2,...,w_l \right] \right) 
\end{equation}
where $h_{plm} \in R^{l\times d}$, $d$ is the hidden dimension of the PLM and $l$ is the number of tokenized tokens. For an entity $e_i$ with $N_{e_i}$ mentions $\left\{ m_{j}^{i} \right\} _{j=1}^{N_{e_i}}$, we apply average pooling to get the \textbf{global} embedding of the entity, $h_{e_i}$.

Attention information from PLM is used to enhance the representations of relation between entity pairs. Specifically, from PLM, we can obtain multi-head attention matrix $A\in R^{H\times l\times l}$, where $A_{k,i,j}$ represents attention score from token $i $ to token $j$ in the $k^{th}$ attention head. Then we extract the attention scores of the head and tail mentions for other mentions from $A$ and average them separately as the entity-level attention score $A_{s}^{E}, A_{o}^{E} \in R^{H\times l}$. Then we compute the localized attention of the entity pair by multiplying their entity-level attention:
    \begin{equation}
A^{(s,o)}=\frac{1}{H}\sum_{i=1}^H{\left( A_{s}^{E}\odot  A_{o}^{E} \right)}
    \end{equation}
where $A^{(s,o)}\in R^{l}$ and $\odot$ denotes element-wise product. Then the \textbf{local} attention-aware entity pair representation is calculated: 
    \begin{equation}
           c^{\left( s,o \right)}=h_{plm}^T\cdot (A^{(s,o)}/1^T\cdot A^{(s,o)})
    \end{equation}
The hybrid representation $z_s \in R^d$ for entity pair $(e_s,e_o)$ is fused with the global and local representation and then transformed by a shared block:
    \begin{equation}
           z_s=\tanh \left( W_s\left[ h_{e_s};c^{(s,o)} \right] +b\right) 
    \end{equation}
where $h_{e_s}$ is the global embedding of the subject and $W_s \in R^{d \times 2d}$ and $b\in R^{2d}$ denote the learnable parameters in the shared block targeted at subject. We obtain $z_o$ for the object in the same way. The representation of the entity pair $(e_s,e_o)$, i.e., the representation of the relation instance is $[z_s;z_o]$. Finally, denoting the set of all instances of relation $r$ in the support set as $S_r$, we obtain the relation prototype $p^r$: 
\begin{equation}
    p^r=\left[ s_{1}^{r},...,s_{\min \left( \left| S_r \right|,\omega \right)}^{r} \right]
\end{equation}
where $s_{i}^{r}$ is the representation of the $i^{th}$ relation instance in $S_r$, and $\omega$ is a hyperparameter that indicates the maximum for selecting candidate relation instances.

\subsection{Transferable Proto-Learner}
\sloppy
To improve the cross-domain transferable capability of the model, we design domain-adaptive Transferable Proto-Learner which can learn deep NOTA features. 
\fussy
Specifically, we design a learnable block to learn the regularized representation of NOTA relation which models the way to compute NOTA prototype to eliminate bias of NOTA in different domains. 
Denoting the set of all instances of NOTA in support documents as $S_{N}$, we obtain the representation of NOTA prototype:
\begin{equation}
p^N=f_{\theta}\left( \left[ s^N_1,...,s^N_\beta \right] \right)
\end{equation}
where $p^N \in \beta\times d$, $f_\theta$ denotes the transferable learnable block with learnable parameters $\theta$ and $\beta$ is a factor that indicates the number of selected instances from $S_N$. In addition, we design double MLPs to model the learnable block, which is plug-and-play for other OOD Detection problems.

\begin{table*}[t]
\centering
\renewcommand{\arraystretch}{1.1}
\scalebox{0.8}{
\begin{tabular}{lcccccccc}
\hline
\multirow{3}{*}{\textbf{Model}} & \multicolumn{4}{c}{\textbf{FREDo}} & \multicolumn{4}{c}{\textbf{ReFREDo}} \\ 
 \cmidrule(lr){2-5} \cmidrule(){6-9} & \multicolumn{2}{c}{\textbf{1-Doc}} & \multicolumn{2}{c}{\textbf{3-Doc}} & \multicolumn{2}{c}{\textbf{1-Doc}} & \multicolumn{2}{c}{\textbf{3-Doc}} \\ \cmidrule(lr){2-3} \cmidrule(r){4-5} \cmidrule(r){6-7} \cmidrule(){8-9} 
& \textbf{In-Domain}         & \textbf{Cross-Domain}         & \textbf{In-Domain }         & \textbf{Cross-Domain}         & \textbf{In-Domain}         & \textbf{Cross-Domain}         & \textbf{In-Domain }          & \textbf{Cross-Domain}          \\ 
\hline
DL-Base  & 0.6  & 1.76  &0.89 & 1.98& 1.38& 1.76
& 1.84
&1.98\\
DL-MNAV &$7.05 \pm 0.18$  & $0.84 \pm 0.16$ & $8.42 \pm 0.64$ &  $0.48 \pm 0.21$ & $12.97\pm0.88$& $ 1.12\pm0.38$
&$12.34\pm0.36$
&$2.28\pm0.19$                \\
DL-MNAV$_{SIE}$ & $7.06 \pm 0.15$ & $1.77 \pm 0.60$ & $6.77 \pm 0.21$ & $2.51 \pm 0.66$& $13.37\pm0.98$&$1.39\pm0.74$
& $12.00\pm0.80$
&$2.92\pm0.41$\\
DL-MNAV$_{SIE+SBN}$ & $1.71 \pm 0.04$& $2.85 \pm 0.12$ & $2.79 \pm 0.24$ & $3.72 \pm 0.14$& $4.59\pm0.30$&$2.84\pm0.24$
& $5.43\pm0.24$
&$3.86\pm0.27$ \\
ATLOP& 6.33& 1.66& 7.08& 1.42  &-&-&-&-\\
HCRP& 6.41& 1.71&6.22& 1.50&-&-         &  -  & - \\
CHAN& 7.91& 1.95& 8.65&2.78&-&-&-&- \\
KDDocRE&   $2.59\pm0.71$&$1.03\pm0.31$&$4.66\pm0.83$
&$2.00\pm0.46$& $4.76\pm0.55$&$2.30\pm0.59$
& $9.02\pm0.64$
&$3.61\pm0.43$\\
RAPL & $8.75\pm0.80$&$3.33\pm0.50$ & \boldmath{$10.67\pm0.77$}
&\underline{$5.35\pm0.72$}& \underline{$15.20\pm0.80$}&\underline{$3.51\pm0.79$}
& \boldmath{$16.35\pm0.60$}&\boldmath{$5.48 \pm 0.63$}\\

\textbf{TPN} &  \boldmath{$9.12 \pm 0.31$}&\underline{$3.98 \pm 0.28$} &     $8.64 \pm 0.44$&$4.48 \pm 0.29$& \boldmath{$15.54 \pm 0.14$}&\boldmath{$4.72 \pm 0.43$}& \underline{$15.73 \pm 0.53$}
&\underline{$5.02 \pm 0.50$}\\
\textbf{TPN(FreeLB)}&\underline{$8.80 \pm 0.62$} & \boldmath{$5.00 \pm 0.32$}
&\underline{$8.68 \pm 0.21$ } &\boldmath{$5.93 \pm 0.44$}&-&-&-&- \\                 \hline
\end{tabular}
}
\caption{Results (Macro-F1 across relation types) on FREDo and ReFREDo. The best score is in bold and the second best is underlined. }
  \label{tab:FREDo_REFREDo}
\end{table*}
\subsection{Dynamic Weighting Calibrator}
Given the $j^{th}$ entity pair in the query document, we can compute its hybrid embedding as $q_j$. Targeted at the entity pair, any relation $r$ from support set is defined as positive class $P^q_j$ when $l_j^r >l_j^N$, otherwise is defined as negative class $N^q_j$, where $l_j^r=max(q_j \cdot p^r)$ and $l_j^N=max(q_j \cdot p^N)$ are the logits of relation $r$ and NOTA. We design a Dynamic Weighting Calibrator to detect relation-specific classification confidence, serving as dynamic weights for different relations. Specifically, the probability of positive and negative classes are computed as: 
\begin{equation}
    P\left( r\right) =\frac{\exp \left( l^r_j \right)}{\exp \left( l^r_j \right) +\exp \left( l^N_j \right)}
\end{equation}

\begin{equation}
    P\left(N \right) =\frac{\exp \left( l^N_j \right)}{\sum_{\hat{r}\epsilon N^q_j\cup \left\{NOTA \right\}}{\exp \left( l_{j}^{\hat{r}} \right)}}
\end{equation}
Furthermore, we use $\left( 1-P\left( r \right) \right) ^{\alpha}$ as dynamic weights to calibrate the NOTA-dominant loss. So the final loss function is computed as:
\begin{small}
    \begin{equation}
        L=\sum_{r\epsilon P^q_j}{\left( 1-P\left( r \right) \right) ^{\alpha}\log \left( P\left( r \right) \right)}+\log \left( P\left(N \right) \right) 
    \end{equation}
\end{small}
where $\alpha$ is a hyper-parameter.

\begin{table}[t]
  \centering
  \renewcommand{\arraystretch}{1.1}
\scalebox{0.65}{
    \begin{tabular}{lcccc}
    \toprule
    
    \multirow{2}{*}{\textbf{Model}}& \multicolumn{2}{c}{\textbf{1-Doc}} & \multicolumn{2}{c}{\textbf{3-Doc}} \\ 
    \cmidrule(l){2-3} \cmidrule(l){4-5}&
    \textbf{In-domain}  & \textbf{Cross-domain}  & \textbf{In-domain}  & \textbf{Cross-domain} \\
    \midrule
 DL-MNAV$_{SIE}$&9.92 &\underline{4.68}& \textbf{27.91}&\underline {5.42}\\
 DL-MNAV$_{SIE+SBN}$&\underline{14.89} &4.15&20.63&4.87\\
 RAPL&13.91& 3.87& 19.99&6.34\\
\textbf{TPN (Ours)}& \boldmath{$ 22.63\pm 1.97$}&\boldmath{$7.41 \pm 0.49$}
& $26.04 \pm 2.74$
&\boldmath{$7.16 \pm 0.68$}\\

    \bottomrule
    
    \end{tabular}
    }
    \caption{Results (Micro-F1 across relation types) on ReFREDo benchmark. The best score is in bold and the second best is underlined.}
  \label{tab:ReFREDo_micro}
\end{table}

\begin{table}[t]
  \centering
    \scalebox{0.8}{
    \begin{tabular}{lcccc}
    \toprule
    \multirow{2}{*}{\textbf{Model}}& \multicolumn{2}{c}{\textbf{1-Doc}} & \multicolumn{2}{c}{\textbf{3-Doc}} \\ 
    \cmidrule(l){2-3} \cmidrule(l){4-5}&
    \textbf{In-domain}  & \textbf{Cross-domain}  & \textbf{In-domain}  & \textbf{Cross-domain} \\
    \midrule
TPN &\textbf{15.54} & \textbf{4.72}&\textbf{15.73}   &\textbf{5.02} \\
  w/o HE &2.91 & 2.07 &2.90  &2.50 \\
  w/o TPL &4.75 &2.46  &4.44  &2.98\\
  w/o DWC & 12.85& 4.63 & 13.24 &5.08\\
\hline
    
    \end{tabular}
    }
    \caption{Ablation study results for three key components on ReFREDo.}
  \label{tab:ablation-refredo}
\end{table}

\subsection{Virtual Adversarial Training}
To alleviate cross-domain generalization performance challenge, we use virtual adversarial training (VAT) to smooth the semantic space and enhance the model's robustness. Specifically, given a text, we add an adversarial normalized perturbation $\xi$ on the word embeddings of PLM while assuming the model prediction should not change after the perturbation. We adopt FreeLB \cite{FreeLB} to perform multiple iterations of PGD to craft adversarial examples and simultaneously accumulate the gradients $\nabla _{\theta}L$ in each iteration. After that, it updates the model parameter $\theta$ at once. By taking a descent step along the averaged gradients at $x+\xi_0,...,x+\xi_{\rho-1}$, we approximately optimize the following objective: 

\begin{small}
\begin{equation}
\underset{\theta}{\min}E\left( z_T,y \right) \sim D\left[ \frac{1}{\rho}\sum_{t=0}^{\rho-1}{\underset{\xi \in S_{V}}{\max}\,\,L\left( \phi_{\theta}\left( x+\xi _t \right) ,z_T \right)} \right] 
\end{equation}
\end{small}

where $\rho$ is the time of PGD iterations, $y=\phi _\theta(x+\xi)$ is the output of the language model, $z_T$ denotes the ground truth, $S_{V}$ denotes the set of the adversarial perturbations $\xi_t$ at the $t^{th}$ step. 
Under the constraint $\left\| \xi \right\| _F\leqslant \epsilon $, the adversarial perturbation $\xi$ is updated in each iteration:

\begin{small}
\begin{equation}
        \xi _t=\varPi _{\left\| \xi \right\| _F\leqslant \epsilon}\left( \xi _{t-1}+\gamma \cdot g_{adv}/\left\| g_{adv} \right\| _F \right) 
    \end{equation}
\end{small}

where $\gamma$ is the step size and the $g_{adv}$ is the gradient of the loss for $\xi$. Compared with PGD, FreeLB achieves more comparable robustness and generalization by taking one descent step on the parameters together with each of the $\rho^{th}$ ascent steps to search more stable position on the perturbation.

\section{Experiments}

\subsection{Benchmarks and Evaluation Metrics}
We conduct extensive experiments on two public FSDLRE benchmarks, FREDo \cite{fredo} and ReFREDo \cite{ReFREDo}. \textbf{FREDo} has been meticulously reconstructed from the amalgamation of the DocRED \cite{docred} and sciERC \cite{sciERC} datasets. Notably, the proportion of NOTA relation in FREDo significantly surpasses other benchmarks, standing at 96.4\%. Training set comprises 62 relation types sourced from Wikipedia, with 16 relation types from the same domain serving as the validation set. For in-domain testing, 16 relation types from Wikipedia are utilized, while cross-domain testing is conducted using 7 relation types extracted from the sciERC corpus. ReFREDo has been reconstructed through a more complete annotation on FREDo. In \textbf{ReFREDo}, the training, validation and in-domain test corpus is replaced as Re-DocRED \cite{docred-revisit2}, but the cross-domain test episodes remain the same as in FREDo. We use Macro-F1 and Micro-F1 as metrics to evaluate our model.

\subsection{Implementation Details}
We use bert-base-cased \cite{bert} from Huggingface's Transformers \cite{transformers} as the encoder and Adamw \cite{adamw} as optimizer with a linear warm-up \cite{warmup} followed by a linear learning rate decay. We train the model for 50k episodes and perform early stopping based on the Macro-F1 on the development set. We take the learning rate as 2e-6. The hyperparameters $\omega$, $\beta$, $\alpha$, $\gamma$, $\varepsilon$ and $\rho$ are set to 10, 10, 1, 0.15, 0.45 and 3. We clip the gradients to a max norm of 1.0 for all tasks. All hyperparameters are tuned on the development set. We report the mean and standard deviation of different metrics by five training trials with different random seeds. We use gradient clipping of 1.0. All models were trained on NVIDIA 3090 GPUs or NVIDIA 4090 GPUs.

\begin{figure}[t]
    \centering
    
    \begin{subfigure}{0.23\textwidth}
        \includegraphics[width=1\linewidth]{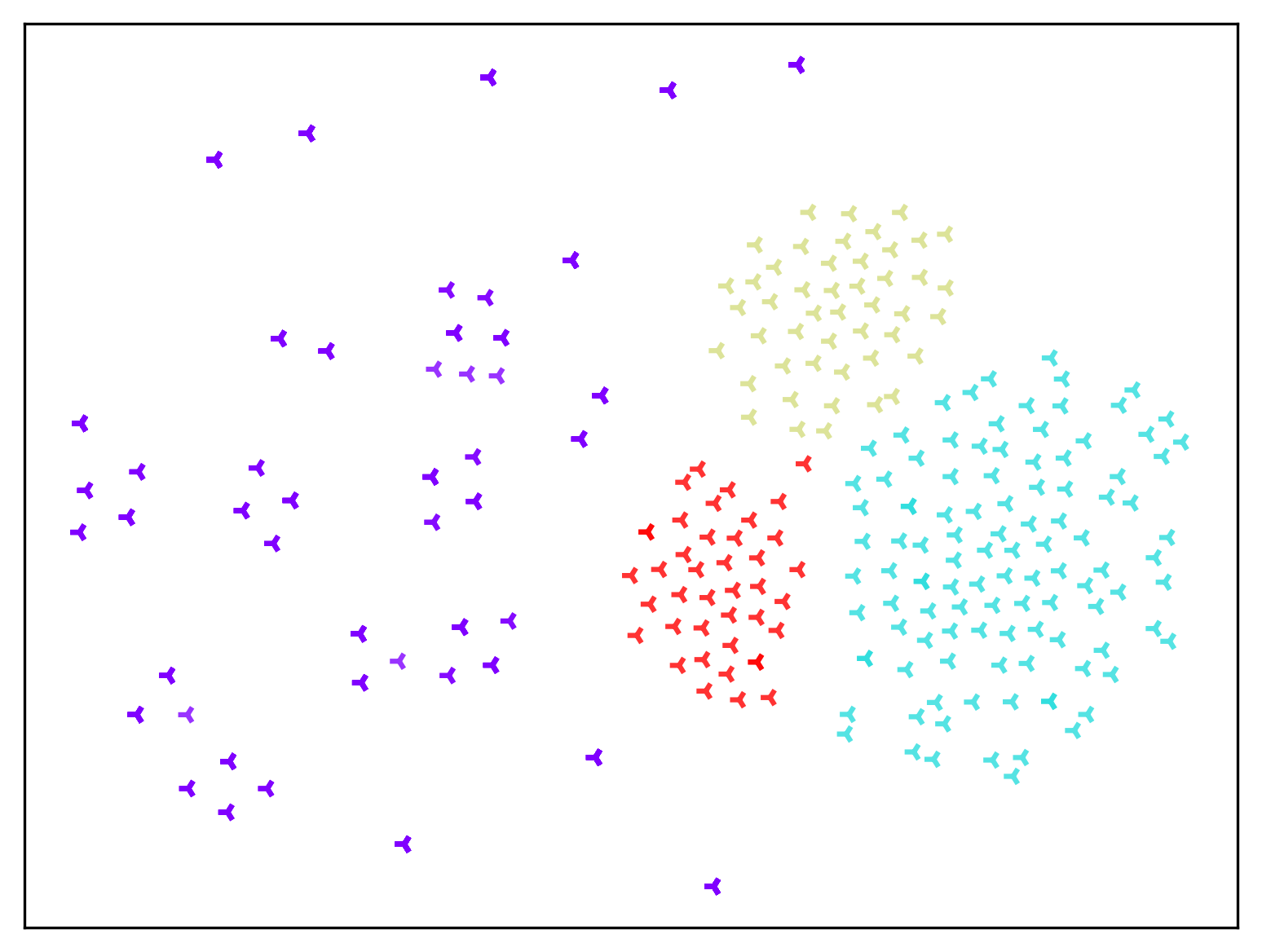}
        \caption*{(a) Proto-DL-MNAV$_{SIE}$}
    \end{subfigure}
    \hfill
    \begin{subfigure}{0.23\textwidth}
        \includegraphics[width=1\linewidth]{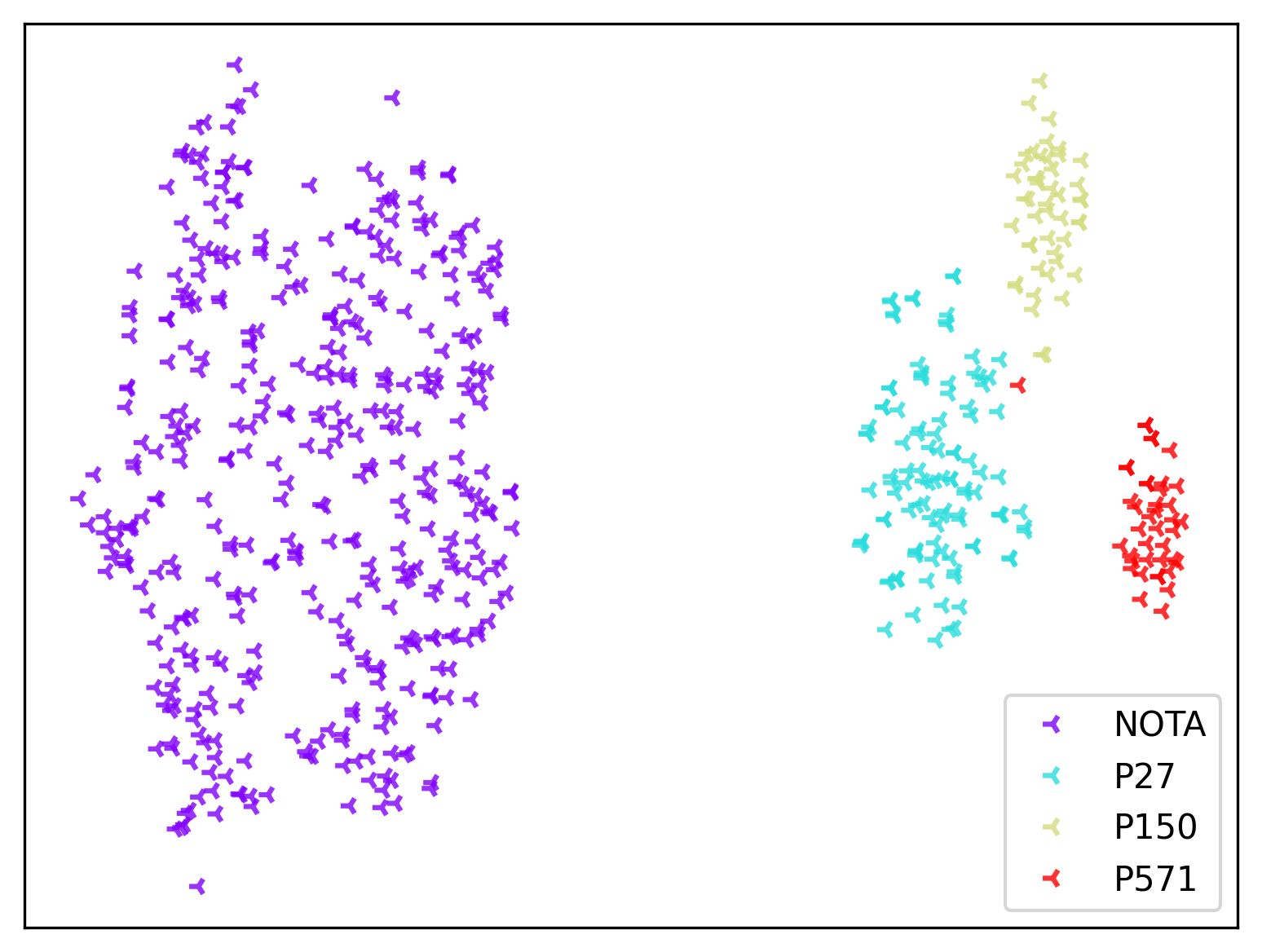}
        \caption*{(b) Proto-TPN}
    \end{subfigure}
    \caption{ Visualization of extracted prototypes for the support set from validation dataset.}
    \label{fig:visualvalid}
\end{figure}

\begin{figure}
    \centering
    \begin{subfigure}{0.23\textwidth}
        \includegraphics[width=\linewidth]{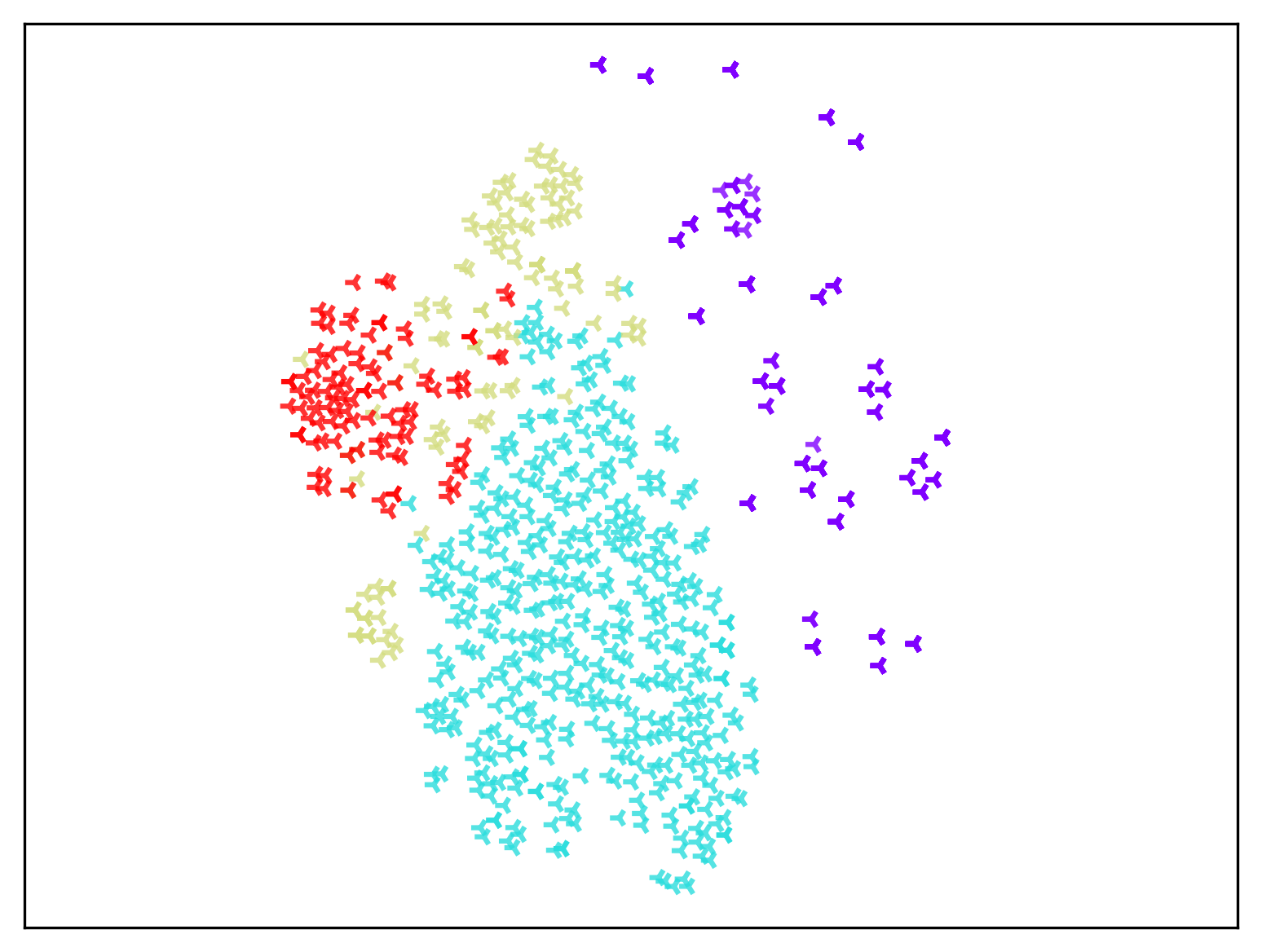}
        \subcaption*{(a) Proto-DL-MNAV$_{SIE}$}
    \end{subfigure}
    \hfill
    \begin{subfigure}{0.23\textwidth}
        \includegraphics[width=\linewidth]{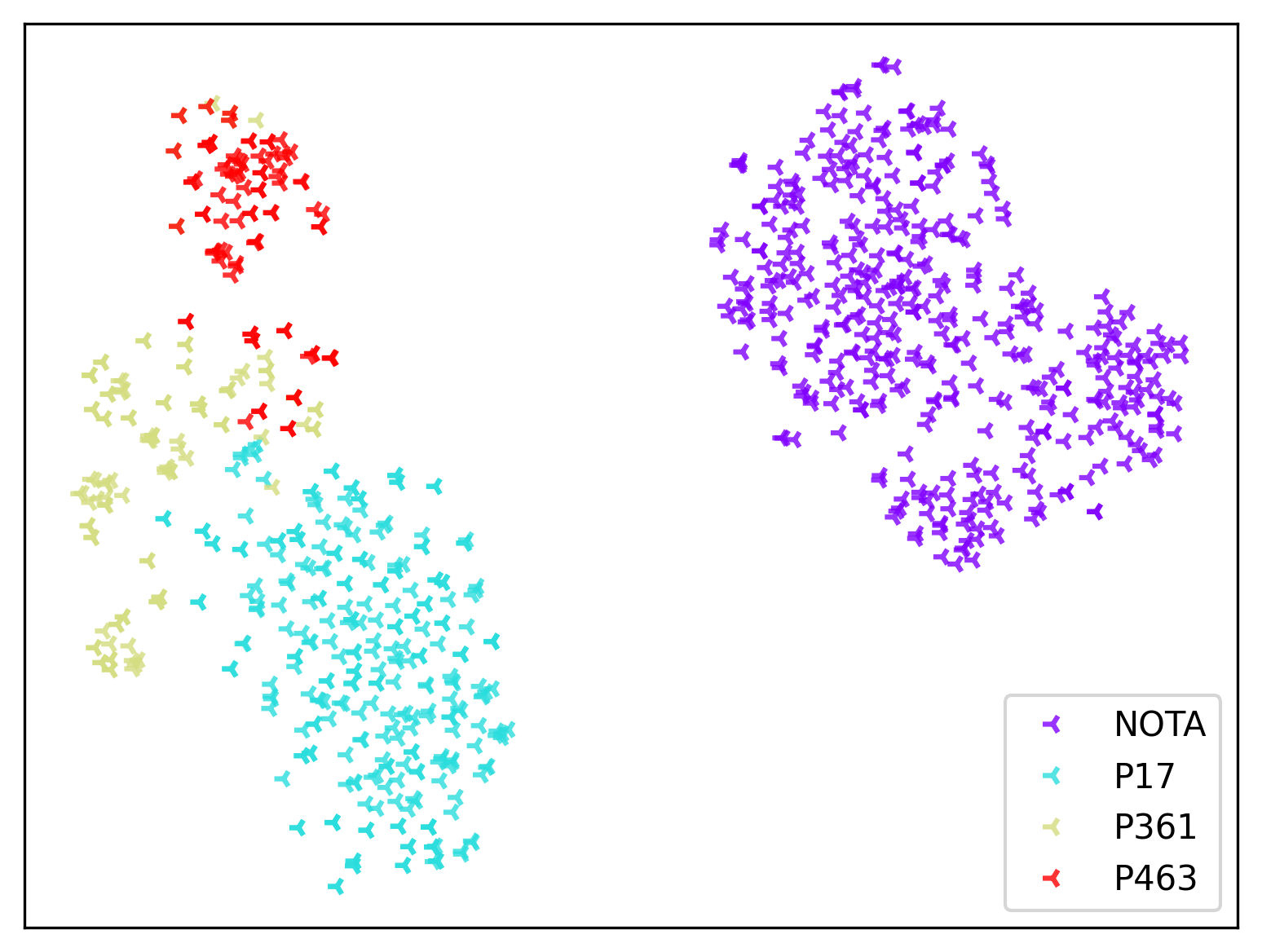}
        \subcaption*{(b) Proto-TPN}
    \end{subfigure}
    \caption{ Visualization of extracted prototypes for the
support set from in-domain testing dataset.}
\label{fig:visualindomain}
\end{figure}

We compare our TPN with a series of baseline methods: \textbf{ATLOP} \cite{atlop} uses Adaptive Thresholding Loss to tackle the multi-class classification. \textbf{HCRP} \cite{hcrp} uses relation description files as clues to capture useful context from sentences. \textbf{CHAN} \cite{chan} designs instance-specific attention to localize the relevant context for each pair of entities. \textbf{KDDocRE} \cite{focal-loss} is the state-of-the-art public-available supervised DLRE method. \textbf{RAPL} \cite{ReFREDo} proposes a relation-aware prototype learning method for FSDLRE to strengthen the relational semantics of prototype representations. \textbf{FREDo} \cite{fredo}: (1) \textbf{DL-Base} only uses the pre-trained language model bert-base-cased \cite{bert} to encode each document and then average the output tokens of each mention of the entity. (2) \textbf{DL-MNAV} applies MNAV \cite{sabo-etal-2021-revisiting} to document-level relation extraction. (3) \textbf{DL-MNAV$_{\textbf{SIE}}$} uses all individual support instances during inference, instead of their average. (4) \textbf{DL-MNAV$_{\textbf{SIE+SBN}}$} uses only the NOTA vectors sampled from the support document and ignores the learned vectors during inference in a new domain.

\subsection{Results}
We denote TPN (FreeLB) as using virtual adversarial training to complement TPN. Due to the limitation of the computation source, we don't evaluate TPN (FreeLB) on ReFREDo benchmark. The main results on FREDo and REFREDo are shown in Table \ref{tab:FREDo_REFREDo} and Table \ref{tab:ReFREDo_micro}. According to the experimental results, we have the following observations: From the metric of Macro-F1: (1) Compared with RAPL, TPN achieves the best average results and stability in the 1-Doc tasks and competitive results in the 3-Doc task for both benchmarks with approximately half the parameter size. 3-Doc task, with more support document supervisory signals for computing the prototype, does not get a better boost in effectiveness, which is likely due to the limitations imposed by the number of selected instances and the simplicity of the learnable block in Transferable Proto-Learner. (2) TPN (FreeLB) obtains the best results in the cross-domain setting, exceeding other methods by a significant margin. This indicates the substantial benefits brought by generative adversarial training in enhancing model robustness. (3) Compared to other methods, TPN exceeds them by a large margin in all tasks, demonstrating the superiority of our method. From the metric of Micro-F1: (4) As shown in Table \ref{tab:ReFREDo_micro}, our approach outperforms all methods significantly.

\begin{figure}[t]
    \centering
    \begin{subfigure}{0.23\textwidth}
        \includegraphics[width=\linewidth]{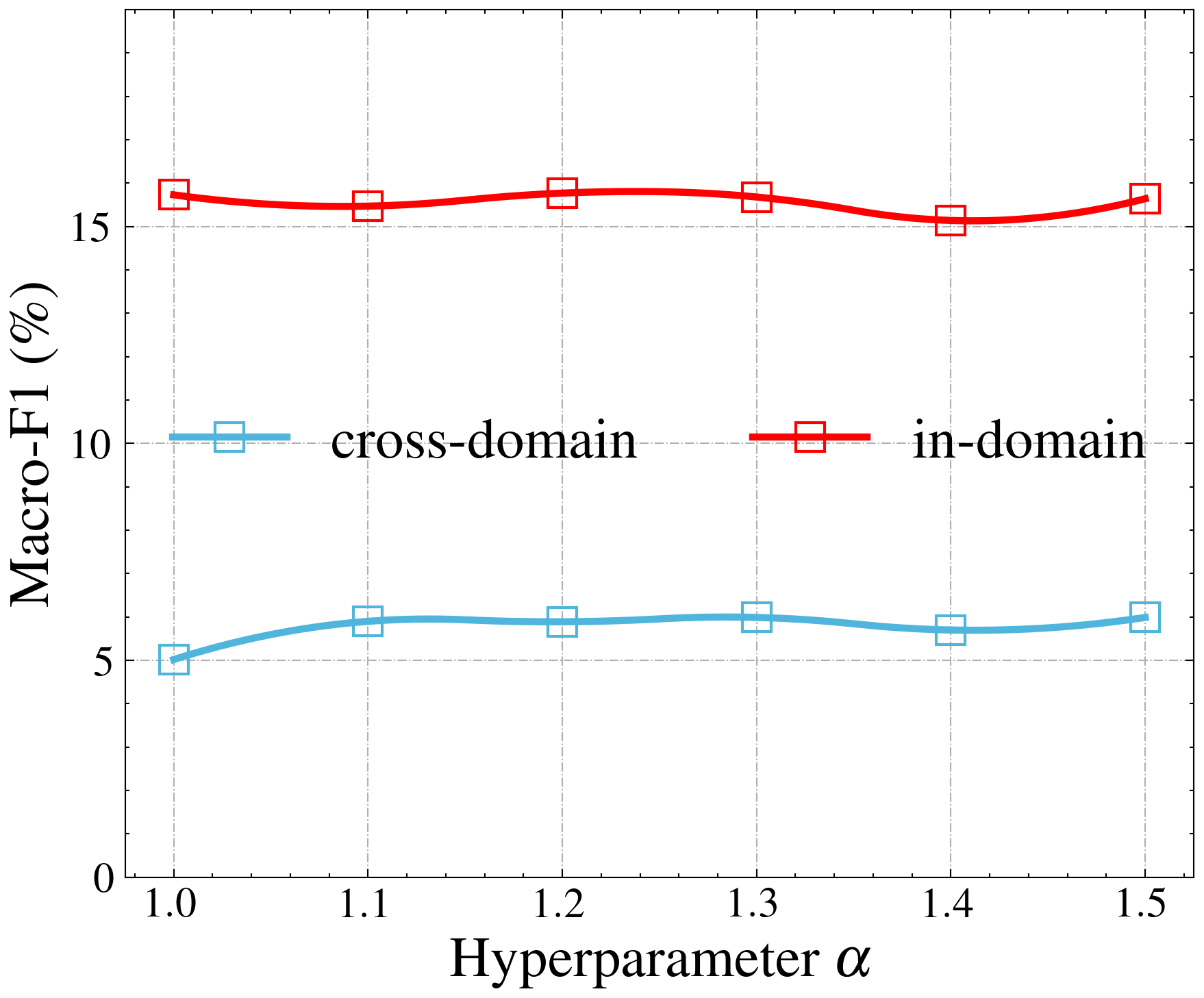}
        \subcaption*{(a) Hyperparameter $\alpha$} 
    \end{subfigure}
    \hfill
    \begin{subfigure}{0.23\textwidth}
        \includegraphics[width=\linewidth]{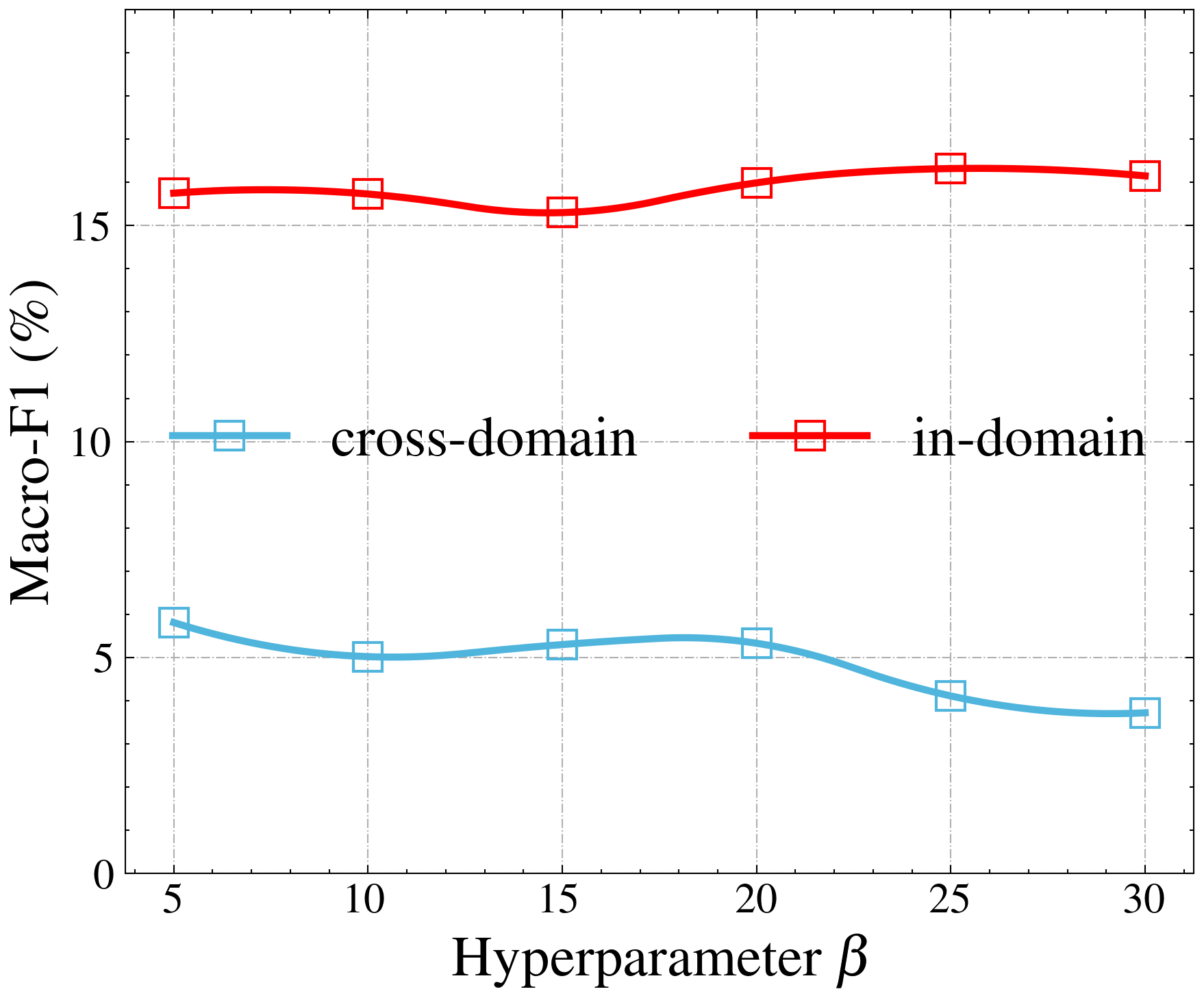}
        \subcaption*{(b) Hyperparameter $\beta$}
    \end{subfigure}
    \caption{Effect of hyperparameter $\alpha$ and $\beta$ under the 3-Doc task setting in ReFREDo.}
    \label{fig:hyper}
\end{figure}
\begin{figure*}[t]
\centering
\includegraphics[width=0.8\linewidth]{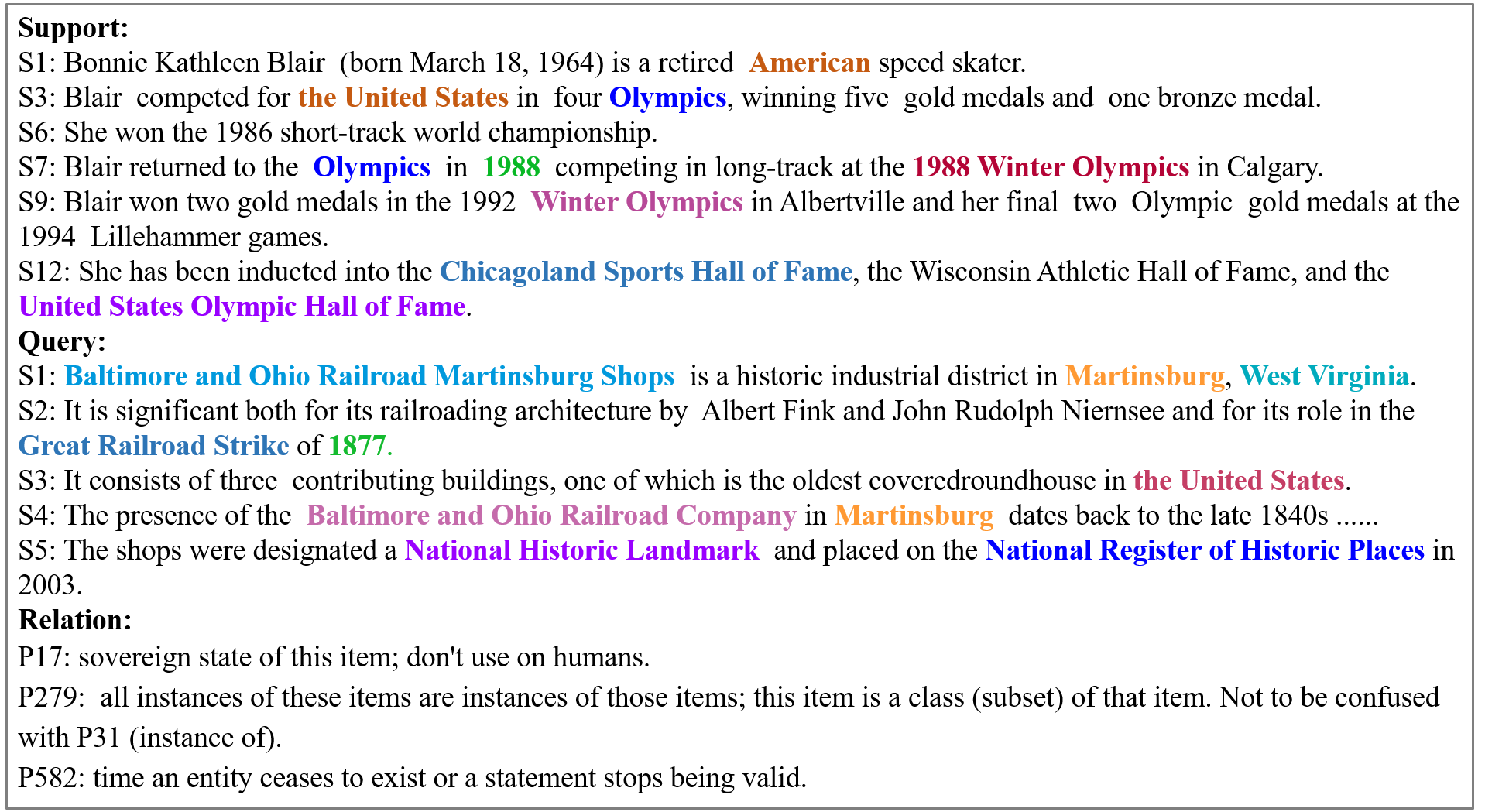}
    \caption{Case study of an in-domain 1-Doc episode in ReFREDo. The positions of entities have been color-coded.}
    \label{fig:case_support_query}
\end{figure*}

\begin{figure*}[t]
\centering
\includegraphics[width=0.8\linewidth]{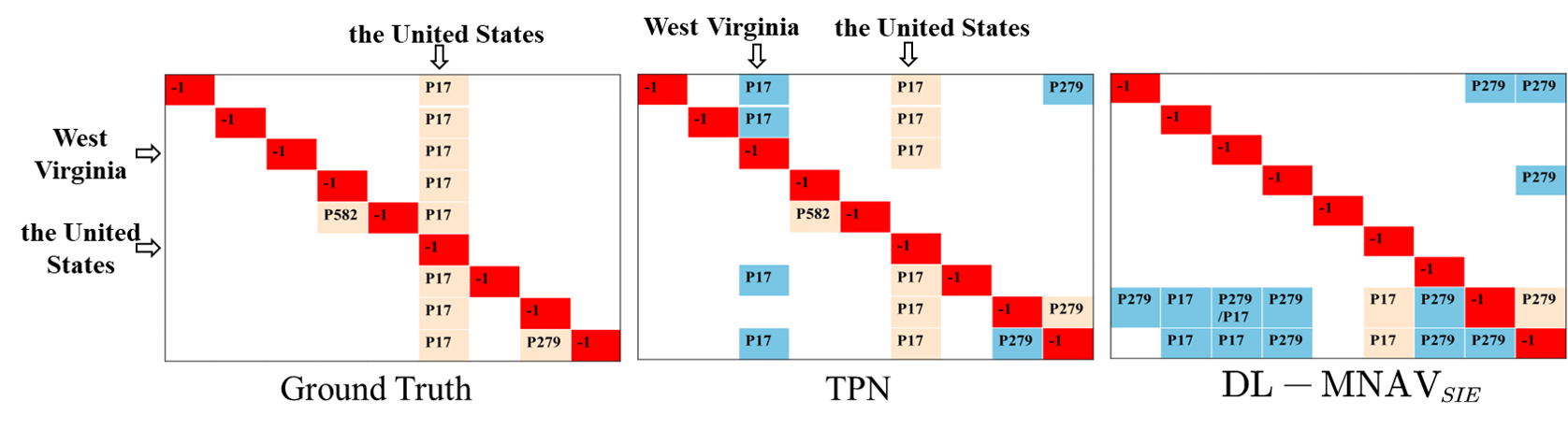}
    \caption{Entity-level Relation Matrix. The indices on the horizontal and vertical axes represent the sequence numbers of entities appearing in the document, while the numbers within the table indicate the relation types between entity pairs.}
    \label{fig:case_pred}
\end{figure*}

\subsection{Ablation Study}
We conduct a thorough ablation study to investigate the effectiveness of three key components in our methods: Hybrid Encoder (HE), Transferable Proto-Learner (TPL), and Dynamic Weighting Calibrator (DWC). From Table \ref{tab:ablation-refredo}, we can observe that all components contribute to model performance. Removal of the HE module results in the most significant performance degradation in all four tasks, which indicates the crucial role of the Hybrid Encoder as the backbone network for encoding complex document information. Furthermore, omitting the TPL module leads to an average decline of 11.04\% in the In-domain and 2.15\% in cross-domain. As a crucial component for adjusting the predicted distribution of positive samples, we argue that DWC will play a more significant role as a class balance predictor when the predictive capability of the model improves.

\section{Analyses and Discussions}
We conduct a comprehensive evaluation and analysis of TPN from various perspectives to provide detailed guidance for future work.

\subsection{Hard VS Single}
\cite{fredo} introduced various strategies to sample training data, including "single" and "hard" methods. These approaches individually signify the presence of either a single relation type or multiple relation types annotated within an episode in the training dataset. However, due to the large number of instances in the NOTA category within the training data, averaging 96.4\%, employing a 'hard' sampling strategy even fails to alter this imbalance. This might explain why previous models did not benefit from this transition (from single to hard). However, from experiments conducted on the in-domain tasks of ReFREDo, we observed a significant and consistent improvement in performance (1-Doc: from $15.54\%$ to $17.13\%$; 3-Doc: from $15.73\%$ to $17.41\%$). This suggests that our approach is adept at sensitively capturing this subtle shift and enables the model to obtain better performance.

\subsection{Latent Space Visualization for Prototype Representation} 
 To illustrate the learned latent space, we visualize the latent variables of the observed characteristics in the 2D plane using t-SNE \cite{t-sne}. We select the top 4 relation types including the NOTA category from the first 40 episodes of the validation and testing datasets of the ReFREDo (3-Doc). In particular, we focus on the support instances embedded in the selected episodes to gain insights into the performance of the TPN and DL-MNAV$_{SIE}$. We can observe that (1) In Fig. \ref{fig:visualvalid}(a), the global NOTA representations of DLMNAV$_{SIE}$ occupy a larger space in the validation data set, which aligns with our previous knowledge that the NOTA space is more extensive, but this effect is not evident in the in-domain dataset from Fig. \ref{fig:visualindomain}(a). (2) From Fig. \ref{fig:visualvalid} and Fig. \ref{fig:visualindomain}, the TPN effectively distinguishes the NOTA relation from the other relations, exhibiting a clear demarcation in the spatial domain. This distinction is notably pronounced in both the validation set and the in-domain testing dataset. In contrast, using DLMNAV$_{SIE}$, the classification boundary between the NOTA relation and other relations appears to be ambiguous. (3) As discussed above, the challenge of constrained and uneven category distribution poses a significant obstacle. Even our method struggles to establish very clear boundaries between known categories, which is a potential area for improvement.
\subsection{Effect of Hyperparameter}
We tune the hyper-parameter $\alpha$ in the Dynamic Weighting Calibrator and $\beta$ in the Transferable Proto-Learner, then evaluate the performance of TPN on the 3-Doc task setting in ReFREDo. Fig. \ref{fig:hyper} shows the performance of TPN with different hyperparameters. 

It is imperative to acknowledge the potential for catastrophic outcomes when $\alpha$ is less than 1, which manifests in significantly amplified 
$(1-P(r_i))^{\alpha}$ value. We observe that optimal performance occurs in the vicinity of 1.1-1.3 for $\alpha$. However, with variations of the hyperparameter, our model exhibits relatively stable performance, which implies robustness in its effectiveness. For $\beta$, when $\beta$ is between 5 and 20, the model exhibits stable performance. However, when $\beta$ exceeds 20, a noticeable performance decrease is observed in the cross-domain setting, while in the in-domain setting there is a slight improvement. This may be attributed to the overfitting of the same domain caused by the excessively large $\beta$.

\subsection{Case Study}
To better demonstrate the performance of our model, we list an example from the in-domian tesing set in Fig. \ref{fig:case_support_query} and Fig. \ref{fig:case_pred} where red block represents that head entity and tail entity are the same, blue block and beige block represent that entity pair is predicted as False Positive and True Positive respectively. DL-MNAV$_{SIE}$ correctly predicts three triples but introduces a considerable number of False Negatives. Our approach predicts the majority of relations including $<$West Virginia, P17, the United States$>$, although mistakenly treating West Virginia (state) as country resulting in four False Negatives for the P17 relation. But from the perspective of the global and local semantics of the document itself, it is difficult to distinguish the subtle difference between West Virginia (state) and country.

\section{Conclusion}
We propose a TPN framework based on Hybrid Encoder, Transferable Proto-Learner, and Dynamic Weighting Calibrator, and use virtual adversarial training to adapt to more complex semantic scenarios targeted at few-shot document-level relation extraction. We conduct an extensive set of experiments to comprehensively showcase the capabilities of the model. Compared with the state-of-the-art methods, our method achieves competitive performance with approximately half the parameter size.

\section*{Acknowledgment}
This work was supported by the National Natural Science
Foundation of China (No. 62276053).


\bibliographystyle{IEEEtran}
\bibliography{main}

\begin{thebibliography}{10}
\providecommand{\url}[1]{#1}
\csname url@samestyle\endcsname
\providecommand{\newblock}{\relax}
\providecommand{\bibinfo}[2]{#2}
\providecommand{\BIBentrySTDinterwordspacing}{\spaceskip=0pt\relax}
\providecommand{\BIBentryALTinterwordstretchfactor}{4}
\providecommand{\BIBentryALTinterwordspacing}{\spaceskip=\fontdimen2\font plus
\BIBentryALTinterwordstretchfactor\fontdimen3\font minus \fontdimen4\font\relax}
\providecommand{\BIBforeignlanguage}[2]{{%
\expandafter\ifx\csname l@#1\endcsname\relax
\typeout{** WARNING: IEEEtran.bst: No hyphenation pattern has been}%
\typeout{** loaded for the language `#1'. Using the pattern for}%
\typeout{** the default language instead.}%
\else
\language=\csname l@#1\endcsname
\fi
#2}}
\providecommand{\BIBdecl}{\relax}
\BIBdecl

\bibitem{docred}
Y.~Yao, D.~Ye, P.~Li, X.~Han, Y.~Lin, Z.~Liu, Z.~Liu, L.~Huang, J.~Zhou, and M.~Sun, ``{D}oc{RED}: A large-scale document-level relation extraction dataset,'' in \emph{Proceedings of the 57th Annual Meeting of the Association for Computational Linguistics}, Florence, Italy, Jul. 2019, pp. 764--777.

\bibitem{atlop}
W.~Zhou, K.~Huang, T.~Ma, and J.~Huang, ``Document-level relation extraction with adaptive thresholding and localized context pooling,'' \emph{Proceedings of the AAAI Conference on Artificial Intelligence}, vol.~35, no.~16, pp. 14\,612--14\,620, May 2021.

\bibitem{dlre1}
S.~Fan, S.~Mo, and J.~Niu, ``Boosting document-level relation extraction by mining and injecting logical rules,'' in \emph{Proceedings of the 2022 Conference on Empirical Methods in Natural Language Processing}, Abu Dhabi, United Arab Emirates, Dec. 2022, pp. 10\,311--10\,323.

\bibitem{meta-learning}
T.~Hospedales, A.~Antoniou, P.~Micaelli, and A.~Storkey, ``Meta-learning in neural networks: A survey,'' \emph{IEEE Transactions on Pattern Analysis and Machine Intelligence}, vol.~44, no.~9, pp. 5149--5169, 2022.

\bibitem{fredo}
N.~Popovic and M.~F{\"a}rber, ``Few-shot document-level relation extraction,'' in \emph{Proceedings of the 2022 Conference of the North American Chapter of the Association for Computational Linguistics: Human Language Technologies}, Seattle, United States, Jul. 2022, pp. 5733--5746.

\bibitem{9413437}
Y.~Wang, R.~Lou, K.~Zhang, M.~Y. Chen, and Y.~Yang, ``More: A metric learning based framework for open-domain relation extraction,'' in \emph{ICASSP 2021 - 2021 IEEE International Conference on Acoustics, Speech and Signal Processing (ICASSP)}, 2021, pp. 7698--7702.

\bibitem{PGD}
\BIBentryALTinterwordspacing
A.~Madry, A.~Makelov, L.~Schmidt, D.~Tsipras, and A.~Vladu, ``Towards deep learning models resistant to adversarial attacks,'' 2019. [Online]. Available: \url{https://arxiv.org/abs/1706.06083}
\BIBentrySTDinterwordspacing

\bibitem{FreeLB}
\BIBentryALTinterwordspacing
C.~Zhu, Y.~Cheng, Z.~Gan, S.~Sun, T.~Goldstein, and J.~Liu, ``Freelb: Enhanced adversarial training for natural language understanding,'' 2020. [Online]. Available: \url{https://arxiv.org/abs/1909.11764}
\BIBentrySTDinterwordspacing

\bibitem{ReFREDo}
S.~Meng, X.~Hu, A.~Liu, S.~Li, F.~Ma, Y.~Yang, and L.~Wen, ``{RAPL}: A relation-aware prototype learning approach for few-shot document-level relation extraction,'' in \emph{Proceedings of the 2023 Conference on Empirical Methods in Natural Language Processing}, Singapore, Dec. 2023, pp. 5208--5226.

\bibitem{re-sentence2}
X.~Chen, N.~Zhang, X.~Xie, S.~Deng, Y.~Yao, C.~Tan, F.~Huang, L.~Si, and H.~Chen, ``Knowprompt: Knowledge-aware prompt-tuning with synergistic optimization for relation extraction,'' in \emph{Proceedings of the ACM Web Conference 2022}, ser. WWW '22, New York, NY, USA, 2022, p. 2778–2788.

\bibitem{fcds}
X.~Zhu, Z.~Kang, and B.~Hui, ``Fcds: Fusing constituency and dependency syntax into document-level relation extraction,'' in \emph{Proceedings of the 2024 Joint International Conference on Computational Linguistics, Language Resources and Evaluation}, 2024.

\bibitem{re-sentence1}
D.~Sui, X.~Zeng, Y.~Chen, K.~Liu, and J.~Zhao, ``Joint entity and relation extraction with set prediction networks,'' \emph{IEEE Transactions on Neural Networks and Learning Systems}, 2023.

\bibitem{Document}
\BIBentryALTinterwordspacing
H.~Liu, Z.~Kang, L.~Zhang, L.~Tian, and F.~Hua, ``Document-level relation extraction with cross-sentence reasoning graph,'' 2023. [Online]. Available: \url{https://arxiv.org/abs/2303.03912}
\BIBentrySTDinterwordspacing

\bibitem{focal-loss}
Q.~Tan, R.~He, L.~Bing, and H.~T. Ng, ``Document-level relation extraction with adaptive focal loss and knowledge distillation,'' in \emph{Findings of the Association for Computational Linguistics: ACL 2022}, Dublin, Ireland, May 2022, pp. 1672--1681.

\bibitem{long-tail}
W.~Wang, B.~Hu, Z.~Peng, M.~Zhong, Z.~Zhang, Z.~Liu, G.~Zhang, and J.~Zhou, ``Garcia: Powering representations of long-tail query with multi–granularity contrastive learning,'' in \emph{2023 IEEE 39th International Conference on Data Engineering (ICDE)}, Los Alamitos, CA, USA, apr 2023, pp. 3182--3195.

\bibitem{tiefake}
Q.~Guo, Z.~Kang, L.~Tian, and Z.~Chen, ``Tiefake: Title-text similarity and emotion-aware fake news detection,'' in \emph{2023 International Joint Conference on Neural Networks (IJCNN)}.\hskip 1em plus 0.5em minus 0.4em\relax IEEE, 2023, pp. 1--7.

\bibitem{ssan}
B.~Xu, Q.~Wang, Y.~Lyu, Y.~Zhu, and Z.~Mao, ``Entity structure within and throughout: Modeling mention dependencies for document-level relation extraction,'' \emph{Proceedings of the AAAI Conference on Artificial Intelligence}, vol.~35, no.~16, pp. 14\,149--14\,157, May 2021.

\bibitem{transfer-learning}
S.~J. Pan and Q.~Yang, ``A survey on transfer learning,'' \emph{IEEE Transactions on Knowledge and Data Engineering}, vol.~22, no.~10, pp. 1345--1359, 2010.

\bibitem{relation-prompt}
Y.~K. Chia, L.~Bing, S.~Poria, and L.~Si, ``{R}elation{P}rompt: Leveraging prompts to generate synthetic data for zero-shot relation triplet extraction,'' in \emph{Findings of the Association for Computational Linguistics: ACL 2022}, Dublin, Ireland, May 2022, pp. 45--57.

\bibitem{transfer-learning-ability}
J.~Yosinski, J.~Clune, Y.~Bengio, and H.~Lipson, ``How transferable are features in deep neural networks?'' in \emph{Advances in Neural Information Processing Systems}, vol.~27, 2014.

\bibitem{few-shot-RE1}
O.~Sainz, O.~Lopez~de Lacalle, G.~Labaka, A.~Barrena, and E.~Agirre, ``Label verbalization and entailment for effective zero and few-shot relation extraction,'' in \emph{Proceedings of the 2021 Conference on Empirical Methods in Natural Language Processing}, Online and Punta Cana, Dominican Republic, Nov. 2021, pp. 1199--1212.

\bibitem{few-shot-RE2}
S.~Yang, Y.~Zhang, G.~Niu, Q.~Zhao, and S.~Pu, ``Entity concept-enhanced few-shot relation extraction,'' in \emph{Proceedings of the 59th Annual Meeting of the Association for Computational Linguistics and the 11th International Joint Conference on Natural Language Processing (Volume 2: Short Papers)}, Online, Aug. 2021, pp. 987--991.

\bibitem{NIPS2017_cb8da676}
J.~Snell, K.~Swersky, and R.~Zemel, ``Prototypical networks for few-shot learning,'' in \emph{Advances in Neural Information Processing Systems}, vol.~30, 2017.

\bibitem{wang-etal-2022-drk}
M.~Wang, J.~Zheng, F.~Cai, T.~Shao, and H.~Chen, ``{DRK}: Discriminative rule-based knowledge for relieving prediction confusions in few-shot relation extraction,'' in \emph{Proceedings of the 29th International Conference on Computational Linguistics}, Gyeongju, Republic of Korea, Oct. 2022, pp. 2129--2140.

\bibitem{label-prompt-dropout}
P.~Zhang and W.~Lu, ``Better few-shot relation extraction with label prompt dropout,'' in \emph{Proceedings of the 2022 Conference on Empirical Methods in Natural Language Processing}, Abu Dhabi, United Arab Emirates, Dec. 2022, pp. 6996--7006.

\bibitem{ood}
S.~Fort, J.~Ren, and B.~Lakshminarayanan, ``Exploring the limits of out-of-distribution detection,'' in \emph{Advances in Neural Information Processing Systems}, vol.~34, 2021, pp. 7068--7081.

\bibitem{sabo-etal-2021-revisiting}
O.~Sabo, Y.~Elazar, Y.~Goldberg, and I.~Dagan, ``Revisiting few-shot relation classification: Evaluation data and classification schemes,'' \emph{Transactions of the Association for Computational Linguistics}, vol.~9, pp. 691--706, 2021.

\bibitem{sciERC}
Y.~Luan, L.~He, M.~Ostendorf, and H.~Hajishirzi, ``Multi-task identification of entities, relations, and coreference for scientific knowledge graph construction,'' in \emph{Proceedings of the 2018 Conference on Empirical Methods in Natural Language Processing}, Brussels, Belgium, Oct.-Nov. 2018, pp. 3219--3232.

\bibitem{docred-revisit2}
Q.~Huang, S.~Hao, Y.~Ye, S.~Zhu, Y.~Feng, and D.~Zhao, ``Does recommend-revise produce reliable annotations? an analysis on missing instances in {D}oc{RED},'' in \emph{Proceedings of the 60th Annual Meeting of the Association for Computational Linguistics (Volume 1: Long Papers)}, Dublin, Ireland, May 2022, pp. 6241--6252.

\bibitem{bert}
J.~Devlin, M.-W. Chang, K.~Lee, and K.~Toutanova, ``{BERT}: Pre-training of deep bidirectional transformers for language understanding,'' in \emph{Proceedings of the 2019 Conference of the North {A}merican Chapter of the Association for Computational Linguistics: Human Language Technologies, Volume 1 (Long and Short Papers)}, Minneapolis, Minnesota, Jun. 2019, pp. 4171--4186.

\bibitem{transformers}
T.~Wolf, L.~Debut, V.~Sanh, J.~Chaumond, C.~Delangue, A.~Moi, P.~Cistac, T.~Rault, R.~Louf, M.~Funtowicz, J.~Davison, S.~Shleifer, P.~von Platen, C.~Ma, Y.~Jernite, J.~Plu, C.~Xu, T.~Le~Scao, S.~Gugger, M.~Drame, Q.~Lhoest, and A.~Rush, ``Transformers: State-of-the-art natural language processing,'' in \emph{Proceedings of the 2020 Conference on Empirical Methods in Natural Language Processing: System Demonstrations}, Online, Oct. 2020, pp. 38--45.

\bibitem{adamw}
J.~Bjorck, K.~Q. Weinberger, and C.~Gomes, ``Understanding decoupled and early weight decay,'' \emph{Proceedings of the AAAI Conference on Artificial Intelligence}, vol.~35, no.~8, pp. 6777--6785, May 2021.

\bibitem{warmup}
\BIBentryALTinterwordspacing
P.~Goyal, P.~Dollár, R.~Girshick, P.~Noordhuis, L.~Wesolowski, A.~Kyrola, A.~Tulloch, Y.~Jia, and K.~He, ``Accurate, large minibatch sgd: Training imagenet in 1 hour,'' 2018. [Online]. Available: \url{https://arxiv.org/abs/1706.02677}
\BIBentrySTDinterwordspacing

\bibitem{hcrp}
J.~Han, B.~Cheng, and W.~Lu, ``Exploring task difficulty for few-shot relation extraction,'' in \emph{Proceedings of the 2021 Conference on Empirical Methods in Natural Language Processing}, Online and Punta Cana, Dominican Republic, Nov. 2021, pp. 2605--2616.

\bibitem{chan}
D.~Wang, S.~Wu, X.~Zhang, and Z.~Feng, ``Multi-relation identification for few-shot document-level relation extraction,'' in \emph{International Conference on Artificial Neural Networks}.\hskip 1em plus 0.5em minus 0.4em\relax Springer, 2023, pp. 52--64.

\bibitem{t-sne}
L.~van~der Maaten and G.~E. Hinton, ``Visualizing data using t-sne,'' \emph{Journal of Machine Learning Research}, vol.~9, pp. 2579--2605, 2008.

\end{thebibliography}

\end{document}